# Masked Generative-Contrastive Representation Learning for Cross-Dataset EEG-Based Emotion Recognition

Huqin Weng[1], Jiayang Huang[1], Yimin Wen, Jie Du, Chi-Man Vong, Chuangquan Chen*

***Abstract*—Self-supervised learning (SSL) shows strong potential for cross-dataset transfer by improving representation learning and model generalizability. However, its application to EEG-based emotion recognition remains largely unexplored. Existing SSL methods struggle to capture the intricate spatiotemporal dependencies of EEG signals under varying channel configurations, extract fine-grained representations resilient to noise, and derive global features that generalize well across subjects. To address these challenges, we propose Masked Generative-Contrastive Representation Learning (MGCRL), a novel SSL framework specifically designed for EEG-based emotion recognition. Built upon a region-aware spatiotemporal encoder, MGCRL integrates generative and contrastive learning to obtain complementary local and global discriminative representations for cross-dataset generalization. MGCRL introduces three key designs: 1) a spatiotemporal encoder that incorporates region-based graph convolution to capture localized spatial and functional relationships, enhancing region-specific feature learning and mitigating the impact of heterogeneous EEG channel layouts across datasets; 2) a generative learning mechanism based on the joint embedding predictive architecture (JEPA) that utilizes masked features to extract noise-robust fine-grained features, improving the model's capability to characterize subtle emotional states; and 3) a contrastive learning strategy that leverages masked and original features to learn temporally stable and cross-subject-invariant embeddings across the same stimuli, boosting emotion discrimination and cross-subject generalization. Under these elaborate designs, MGCRL exhibits a remarkable ability to learn universal representations from unlabeled data. Extensive experiments involving pre-training on the large FACED dataset and fine-tuning on multiple SEED-series datasets demonstrate that MGCRL consistently delivers stable and superior results, outperforming competitive baseline methods in cross-subject emotion recognition tasks.**

* Corresponding author.

1 These authors contributed equally to this work and should be considered co-first authors.

This work was supported in part by the National Natural Science Foundation of China under Grant 62201402, in part by the Guangdong Basic and Applied Basic Research Foundation under Grant 2023A1515011978, and in part by the Guangdong Overseas Distinguished Teacher Project under Grant MS202600204.

Huqin Weng, Jiayang Huang, Yimin Wen and Chuangquan Chen are with the School of Electronics and Information Engineering, Wuyi University, Jiangmen 529020, China (e-mail: wenghuqin@163.com; huangjiayang96@163.com;ymwen2004@aliyun.com;chenchuangquan87@163.com ).

Jie Du is with the School of Biomedical Engineering, Shenzhen University Medical School, Shenzhen University, Shenzhen 518060, China (e-mail: dujie@szu.edu.cn).

Chi-Man Vong is with the Department of Computer and Information Science, University of Macau, (e-mail: cmvong@um.edu.mo).

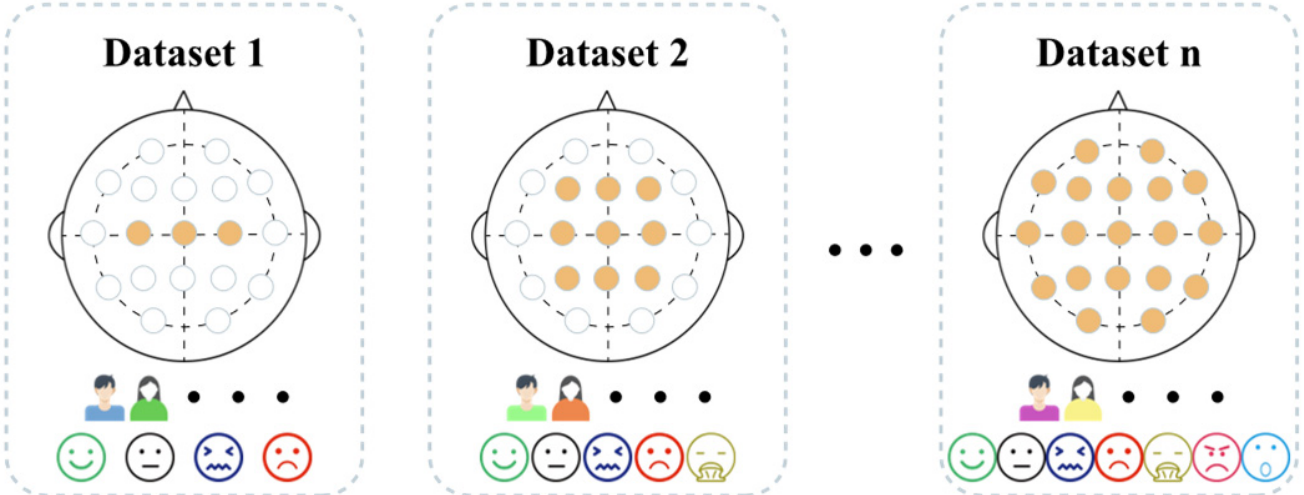

Fig. 1. Illustration of heterogeneities in the cross-dataset problem. These include device heterogeneity, label heterogeneity, and channel heterogeneity.

## I. Introduction

Affective brain-computer interfaces (aBCIs) are emerging technologies that allow machines to monitor, interpret, and modulate human emotional states to support cognition, communication, and decision-making [1], [2]. Researchers have explored multiple neural recording modalities for building aBCIs, including stereoelectroencephalography (SEEG), functional magnetic resonance imaging (fMRI), functional near-infrared spectroscopy (fNIRS), and electroencephalography (EEG). Among these options, EEG has received increasing attention owing to its noninvasiveness, high temporal resolution, and practical ease of use [3]. These strengths, together with the central role of emotion recognition in aBCIs, have driven increasing efforts to develop reliable and accurate EEG-based emotion recognition models [4], [5], [6], [7], [8], [9].

With advances in deep learning, substantial progress has been made in automatically extracting informative features from EEG signals, leading to improved emotion recognition performance [10], [11]. However, most existing approaches are built upon supervised learning, which requires large amounts of labeled data for effective model training. Furthermore, models trained on specific datasets often exhibit limited generalization when applied to different recording conditions or channel configurations. These limitations underscore the need for cross-dataset emotion recognition methods that can learn transferable representations while reducing dependence on labeled data.

EEG-based cross-dataset learning is inherently challenging due to multiple sources of heterogeneity across datasets, including device differences, label incompatibility, and channel mismatches, as shown in Fig. 1. Such discrepancies cause significant inconsistencies in extracted features and hinder model generalization [12]. Recently, self-supervised learning (SSL) has offered a promising solution by exploiting intrinsic structures in unlabeled data to learn robust and transferable representations invariant to these variations [13]. Despite this progress, existing SSL methods applied to EEG-based cross-dataset emotion recognition still face several limitations:

1) Limited spatiotemporal modeling under varying channel configurations. Most existing methods [14], [15] primarily focus on temporal dynamics and fail to fully consider spatial dependencies due to the cross-dataset channel variations, thereby limiting the capture of inter-regional interactions critical for emotion representation.

2) Insufficient fine-grained representations resilient to noise. Most generative learning methods [16], [17], [18] reconstruct signal-level (e.g., raw signals or precomputed Fourier spectrum) representations, resulting in extracted representations that may be susceptible to noise [19], thereby constraining the ability to extract fine-grained representations that preserve subtle and discriminative neural patterns related to different emotional states.

3) Inadequate subject-invariant global representations. Most methods [19], [20] fail to fully account for individual differences and cross-subject similarities in emotional states, which in turn restricts generalization and robustness in cross-subject emotion recognition.

To address these challenges, we propose Masked Generative-Contrastive Representation Learning (MGCRL), a novel SSL framework specifically designed for EEG-based emotion recognition. Specifically, MGCRL designs a region-aware spatiotemporal encoder to capture localized spatial and functional relationships, enhancing region-specific feature learning and mitigating the impact of varying EEG channel configurations across datasets. Additionally, generative learning based on a Joint Embedding Predictive Architecture (JEPA) is adopted to facilitate the learning of noise-robust and fine-grained spatiotemporal representations. Here, "generative" refers to latent representation prediction, which is expected to reduce sensitivity to signal-level noise by encouraging the model to learn abstract latent patterns rather than reconstructing signal-level details [19]. In addition, contrastive learning with masked features is introduced to extract temporally stable and cross-subject-invariant feature representations across the same stimuli. With these carefully designed components, MGCRL demonstrates a remarkable ability to learn universal representations from unlabeled EEG data.

To the best of our knowledge, this is the first attempt to leverage both masked generative learning and contrastive learning for EEG-based emotion recognition directly on raw signals. Unlike conventional multi-task learning schemes that simply combine multiple objectives, MGCRL integrates contrastive and generative paradigms in a mutually complementary manner. Specifically, we ingeniously reuse the masking mechanism, enabling it to serve dual roles: contextual reconstruction in the generative task and data augmentation in the contrastive task. This enables the two learning objectives to mutually constrain each other and be jointly optimized, thereby achieving a structural unification of the two paradigms. Through this design, MGCRL fully exploits the strengths of both learning paradigms while effectively mitigating their respective limitations when applied independently.

The main contributions of this study can be summarized in four aspects:

1) We propose MGCRL, a novel self-supervised learning framework for EEG-based emotion recognition, which leverages a shared masking mechanism to enable generative and contrastive learning to mutually constrain each other and be jointly optimized, thereby capturing both fine-grained and global representations and enhancing the model's ability to characterize subtle and subject-invariant emotional states.

2) We design a region-aware spatiotemporal encoder based on graph convolution, which constructs an adaptive adjacency matrix to jointly model the spatial structure within brain regions and sample-specific functional dependencies, enhancing feature learning under varying EEG channel configurations.

3) We propose a masked dynamic contrastive learning strategy that exploits temporal stability and cross-subject emotional similarity, with masked features as a data augmentation view. This design enables adaptive modeling and iterative refinement of intra- and inter-sample structural relationships, boosting discriminative capability and generalization.

4) We demonstrate that MGCRL achieves robust cross-subject representation learning, outperforming competitive baselines across multiple EEG-based emotion recognition datasets. Notably, the experiments on the SEED-IV, SEED-V, and SEED-VII datasets are conducted under identical hyperparameter configurations, which further underscores the stability and robustness of MGCRL.

The organization of this paper is structured as follows: Section II provides a brief review of related work. Section III details the proposed method. Section IV conducts a comprehensive evaluation of the proposed model. Finally, Section V concludes the paper.

## II. Related Work

This section reviews several related studies to highlight how the proposed MGCRL differs from existing approaches in EEG-based emotion recognition and self-supervised representation learning.

### *A. Cross-Dataset EEG-Based Emotion Recognition*

Most EEG-based emotion recognition studies focus on single-dataset scenarios, while cross-dataset generalization remains challenging due to individual differences, environmental variations, and heterogeneous data acquisition protocols. Bethge et al. [21] adopted an adversarial learning framework to learn domain-invariant feature representations, but only retained 14 common channels during preprocessing. Yuan et al. [22] standardized the number of channels and sampling rates across different datasets, using only 12 common channels for modeling. Although such channel selection ensures cross-dataset consistency, it inevitably leads to the loss of spatial structural information and discards potentially useful information contained in the omitted channels. Gu et al. [23] proposed a dual-branch structure to exploit both shared and dataset-specific channels; nevertheless, it requires consistent label sets, restricting its applicability. In contrast, MGCRL offers a flexible solution capable of handling heterogeneous EEG datasets with varying channels

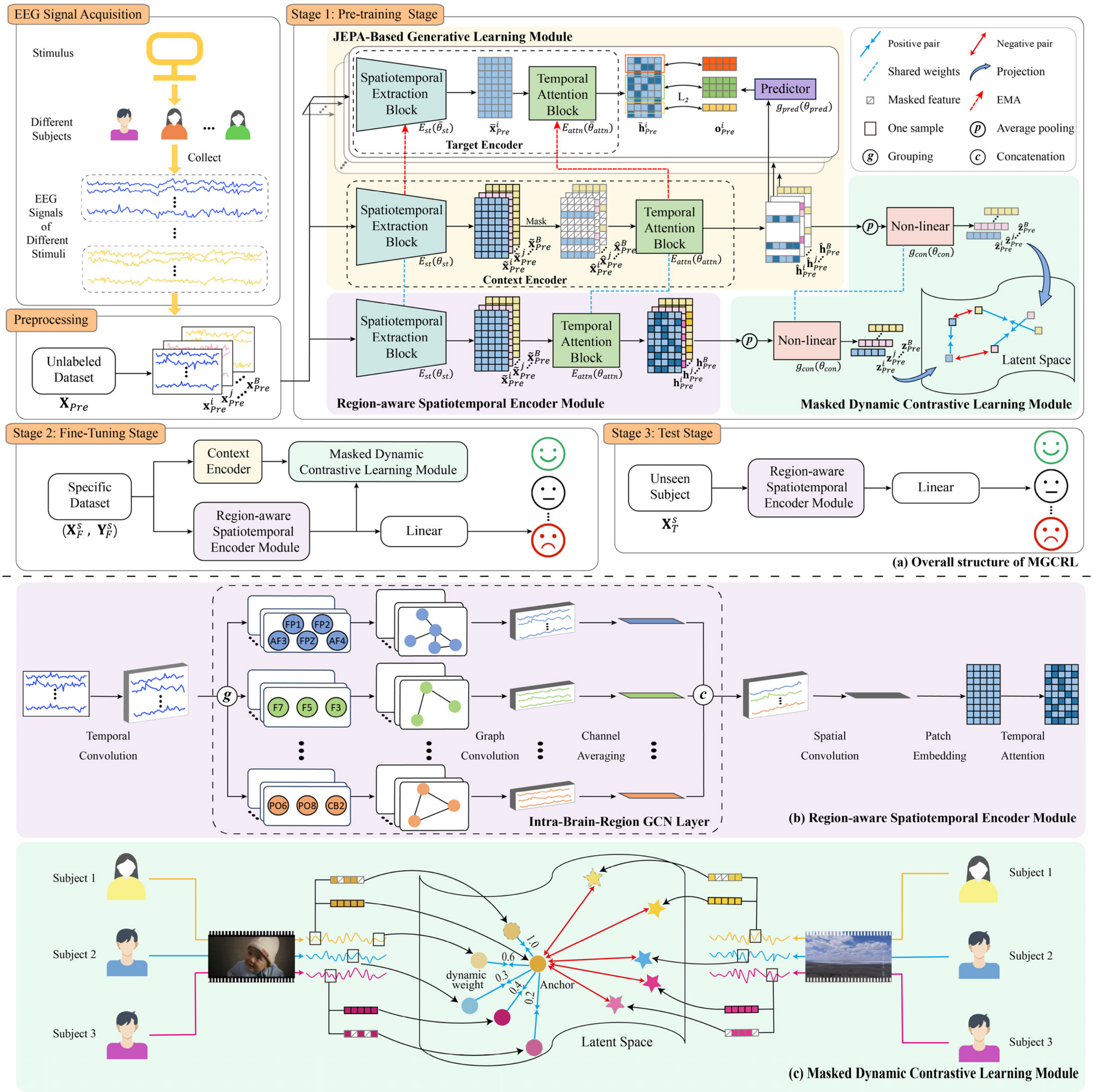

Fig. 2. The overall structure of MGCRL. (a) MGCRL consists of three stages: pre-training, fine-tuning, and testing. During pre-training, the model jointly optimizes JEPA-based generative and masked dynamic contrastive learning module. The resulting model is fine-tuned on a specific dataset and evaluated on an unseen subject to ensure cross-subject generalization. (b) The Region-aware Spatiotemporal Encoder Module. (c) Masked Dynamic Contrastive Learning Module.

and label sets, enabling more robust cross-dataset emotion recognition.

## *B. Generative Self-supervised Learning*

The Masked Autoencoder (MAE), proposed by He et al. [24], serves as a representative method of generative self-supervised learning, achieving efficient representation learning in computer vision through the reconstruction of image patches. Building on this idea, Li et al. [17] adapted the MAE mechanism to EEG-based emotion recognition and proposed the Multi-view Spectral-Spatial-Temporal Masked Autoencoder (MV-SSTMA) to capture the multi-dimensional features of EEG signals. Subsequently, Jiang et al. [16] introduced the Large Brain Model (LaBraM), which decodes both the Fourier spectrum and phase to more comprehensively recover the original signal characteristics. Although these methods differ in the form of reconstruction targets, they all rely on explicit reconstruction of the original input. Such signal-level reconstruction inherently has limitations: the model must expend considerable computational resources to restore low-level details that may contain noise or be irrelevant to the task, thereby increasing training complexity and data requirements. In contrast, I-JEPA [25] adopts abstract prediction targets to avoid overfitting to irrelevant details in pixel-level reconstruction. However, it does not account for the unique spatiotemporal characteristics of EEG signals, which often leads to suboptimal representations when applied to EEG-based emotion recognition. MGCRL addresses this by employing a region-aware spatiotemporal encoder to model localized spatial and functional relationships.

## *C. Contrastive Self-supervised Learning*

Contrastive self-supervised learning aims to learn discriminative representations by encouraging similar samples to cluster together while separating dissimilar samples in the latent space. Existing contrastive methods can be roughly divided into intra-sample and inter-sample methods. In intra-sample contrastive learning, Chen et al. [26] proposed SimCLR, which learns robust visual features by contrasting the consistency of representations of the same sample under different augmentations. Mohsenvand et al. [20] introduced contrastive learning into EEG modeling, enhancing cross-task shared EEG representations by contrasting multiple augmented views. Zhang et al. [15] leveraged the time-frequency consistency principle to enhance the representation transferability of time-series features by contrasting temporal and spectral views of the same sample. While these methods enforce consistency across augmented views within a sample, they do not explicitly model the potential semantic relationships between samples. Inter-sample contrastive learning takes the opposite perspective. Shen et al. [27] proposed CLISA, which aligns features across subjects through contrastive learning to enhance cross-subject emotion recognition. Chang et al. [28] proposed a multi-scale contrastive sampling strategy that constructs different contrasts at the emotional and stimulus levels to maintain the specificity of emotions while reducing individual differences. However, these approaches primarily focus on relationships between samples and overlook intra-sample consistency. Overall, most existing works emphasize either intra-sample or inter-sample consistency separately, with limited efforts toward integrating both with a unified framework. MGCRL addresses this gap by enforcing intra-sample consistency through a masking strategy, while simultaneously modeling inter-sample relationships based on temporal stability and cross-subject emotional similarity under a shared stimulus.

# III. Methods

## *A. Formulation*

To formally define the learning problem, each EEG sample is represented as a unified spatial-temporal tensor. Specifically, the signals are projected onto a 9×9 spatial graph structure based on the standard electrode placement map [29], enabling brain-region segmentation and reducing feature inconsistencies caused by differences in EEG channel configurations between the pre-training and fine-tuning stages. Missing electrode positions are filled using zero-padding. These padded values are used solely for spatial alignment and are not involved in any graph computation or feature aggregation.

The pre-training dataset consists of a large set of unlabeled samples collected from multiple subjects, which is represented as: $\mathbf{X}_{Pre} = \{\mathbf{x}_{Pre}^{(i)}\}_{i=1}^{N_{Pre}} \in \mathbb{R}^{N_{Pre}\times 9\times 9\times T}$ , where $N_{Pre}$ denotes the number of samples and $T$ represents the number of time points per sample. The fine-tuning dataset contains a limited number of labeled samples from a specific dataset $s$, represented as $\mathbf{X}_F^s = \{\mathbf{x}_F^{(i)}\}_{i=1}^{N_F} \in \mathbb{R}^{N_F\times 9\times 9\times T}$ and $\mathbf{Y}_F^s = \{\mathbf{y}_F^{(i)}\}_{i=1}^{N_F}$, where $N_F$ is the number of samples. The test dataset contains samples from an unseen subject whose acquisition device and classification task match those of the fine-tuning dataset $s$, denoted as $\mathbf{X}_T^s = \{\mathbf{x}_T^{(i)}\}_{i=1}^{N_T} \in \mathbb{R}^{N_T\times 9\times 9\times T}$, where $N_T$ is the number of samples.

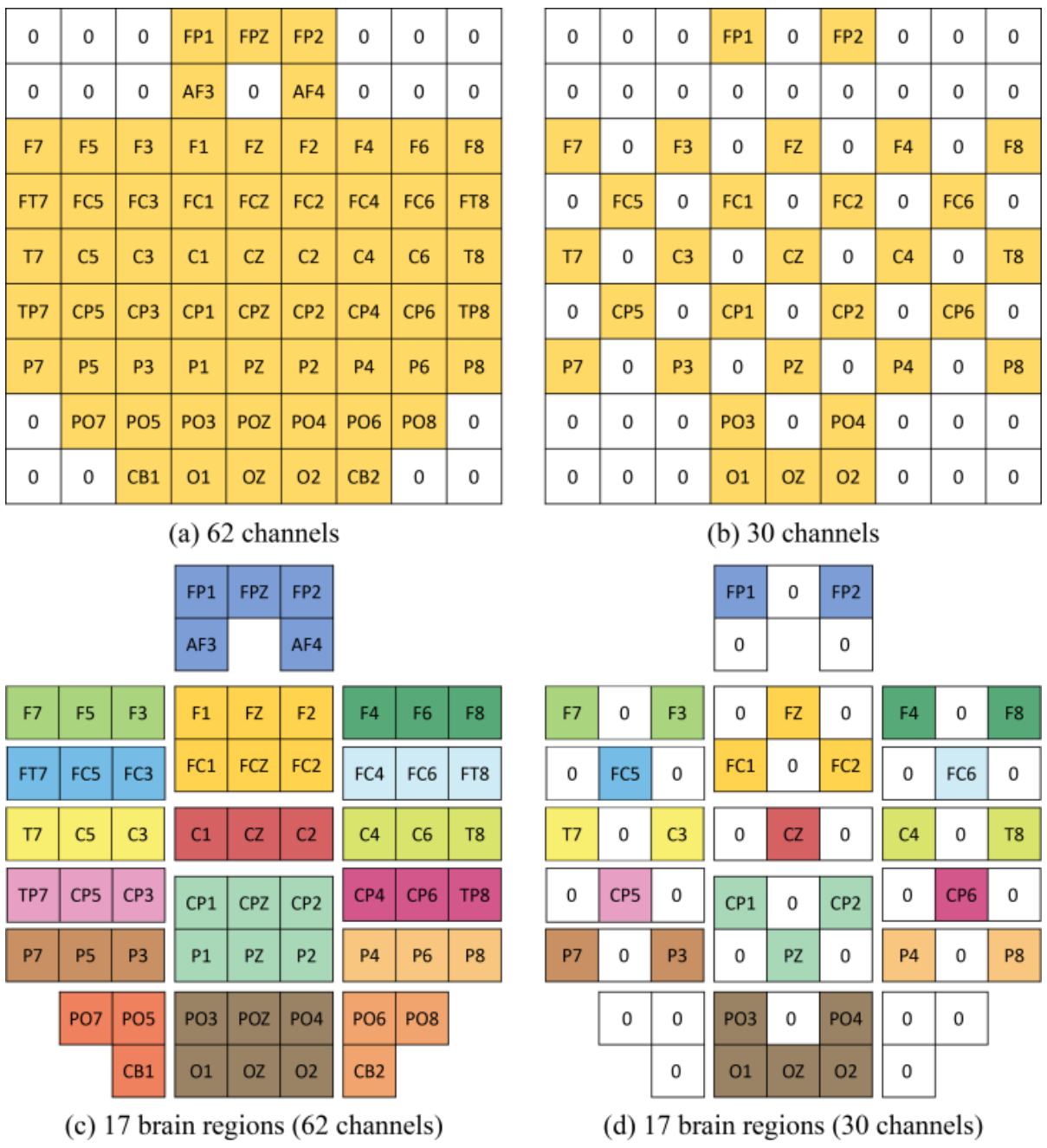

Fig. 3. EEG Channel Mapping (9×9 Grid vs. Brain Region Division), where 0 represents missing channels. (a) Dataset A: 62 channels. (b) Dataset B: 30 channels. (c) Dataset A: 17 brain regions (62 channels). (d) Dataset B: 17 brain regions (30 channels).

## *B. Overview*

As illustrated in Fig. 2(a), the MGCRL framework consists of three stages: pre-training, fine-tuning, and testing. In the pre-training stage, unlabeled EEG data from a large-scale dataset $\mathbf{X}_{Pre}$ are first fed into the region-aware spatiotemporal encoder module (Fig. 2(b)), which simultaneously extracts temporal dynamics, local structural dependencies among brain regions, and global spatial information. A JEPA-based generative learning module is then applied to obtain fine-grained representations, while the masked dynamic contrastive learning module (Fig. 2(c)) captures the temporal stability and cross-subject-invariant emotional features. During the fine-tuning stage, the model is adapted using a limited set of labeled data ($\mathbf{X}_F^s$ , $\mathbf{Y}_F^s$) from a specific dataset $s$. To enhance robustness against individual variability and noisy EEG patterns, the contrastive module is retained in the fine-tuning stage. Finally, in the testing stage, the fine-tuned model is applied to the EEG samples $\mathbf{X}_T^s$ from an unseen subject, enabling subject-independent emotion recognition.

### C. Generalized Pre-training

The pre-training stage is designed to learn generalizable spatiotemporal representations from a large set of unlabeled EEG samples. This stage relies on three specialized modules, which are detailed below.

*1) Region-aware Spatiotemporal Encoder Module:* We design a region-aware spatiotemporal encoder module that jointly models temporal dynamics, local structural dependencies, and global spatial relationships of EEG signals. The region-aware spatiotemporal encoder module consists of a spatiotemporal extraction block $E_{st}$ and a temporal attention block $E_{attn}$ with the parameters $\theta_{st}$ and $\theta_{attn}$, respectively.

The spatiotemporal extraction block begins by applying a $1 \times 5$ temporal convolution (with $K$ kernels) to the input EEG sample $\mathbf{x}_{Pre} \in \mathbb{R}^{9\times 9\times T}$ to learn dynamic-frequency representations [30], [31]. The resulting feature map is denoted as $\check{\mathbf{x}}_{Pre} \in \mathbb{R}^{K\times 9\times 9\times t}$, where $t$ denotes the reduced temporal dimension. As this temporal convolution is a linear operation without bias or activation, the positions corresponding to invalid channels remain zero in $\check{\mathbf{x}}_{Pre}$.

Next, we introduce an intra-brain-region graph convolutional (IBR-GCN) layer that operates on $\check{\mathbf{x}}_{Pre}$ to capture the structural information within specific brain regions. Specifically, we first partition the brain areas following the scheme in [32], as illustrated in Fig. 3. Based on this partitioning, the feature representation $\check{\mathbf{x}}_{Pre}$ is divided into different subsets $\{\check{\mathbf{x}}_{Pre}^1, \check{\mathbf{x}}_{Pre}^2, \dots, \check{\mathbf{x}}_{Pre}^M\}$, where each subset only contains the channels corresponding to the respective brain region. Here, $M$ denotes the number of brain regions ( $M = 17$ in this study, see Fig. 3). The subset of features for brain region $m$ $(m = 1,2,\dots,M)$ is represented as $\check{\mathbf{x}}_{Pre}^m \in \mathbb{R}^{K\times C_m\times t}$, where $C_m$ denotes the number of channels in brain region $m$. To model the spatial relationships among channels within each brain region, a GCN layer is applied to each region. Since the model contains $K$ views and $M$ brain regions, a total of $K \times M$ graph convolution operations are performed. For clarity, we introduce a variable $\check{\mathbf{x}}_k^m \in \mathbb{R}^{C_m\times t}$ to represent the feature of a single brain region under one specific view. The graph convolutional network is then applied to these features as follows. Each EEG channel within the region is treated as a graph node, with its temporal feature vector serving as the node feature. For each sample, a dynamic graph structure $G = (\mathbf{A}_k^m, \check{\mathbf{x}}_k^m)$ is constructed, where the adjacency matrix is defined as:

$$\mathbf{A}_k^m = \mathrm{MASK}\left(\mathrm{ReLU}\left(\mathbf{A}_{dyn,k}^m \circ \left(\mathbf{A}_{phy}^m + \Delta_k^m\right)\right)\right). \tag{1}$$

Here, the physical adjacency matrix $\mathbf{A}_{phy}^m \in \mathbb{R}^{C_m\times C_m}$ reflects the spatial arrangement of electrodes within each brain region. By defining connectivity based on the inverse square of the spatial distance between channels [33], it incorporates prior physiological knowledge that nearby electrodes tend to record more related neural activity. $\Delta_k^m \in \mathbb{R}^{C_m\times C_m}$ is a symmetric and trainable matrix used to adaptively refine the physical adjacency relationships. In contrast, the dynamic functional adjacency matrix $\mathbf{A}_{dyn,k}^m \in \mathbb{R}^{C_m\times C_m}$ is computed from the feature representation $\check{\mathbf{x}}_k^m$ of each sample, i.e., $\mathbf{A}_{dyn,k}^m = \check{\mathbf{x}}_k^m(\check{\mathbf{x}}_k^m)^T$, capturing sample-specific functional dependencies among channels. The element-wise multiplication $\circ$ allows the graph to reflect both physical and functional correlation constraints. The ReLU function ensures that the resulting adjacency values are non-negative, while the MASK operator removes the influence of invalid channels. Specifically, if the feature of channel $i$ is invalid, the MASK operator sets the $i$-th row and the $i$-th column to zero, effectively eliminating all connections associated with that channel.

We normalize the adjacency matrix as follows:

$$\tilde{\mathbf{A}}_k^m = (\mathbf{D}_k^m)^{-\frac{1}{2}}(\mathbf{A}_k^m + \mathbf{I}^m)(\mathbf{D}_k^m)^{-\frac{1}{2}}, \tag{2}$$

where $\mathbf{I}^m \in \mathbb{R}^{C_m\times C_m}$ is the identity matrix, and $\mathbf{D}_k^m = \mathrm{diag}(d_1, d_2, \dots, d_{C_m})$ is the degree matrix of the self-loop-augmented adjacency matrix $\mathbf{A}_k^m + \mathbf{I}^m$ , with $d_i = \sum_{j=1}^{C_m}(\mathbf{A}_k^m + \mathbf{I}^m)(i,j)$ . After constructing the normalized adjacency matrix, the graph convolution layer can be formulated as:

$$\check{\mathbf{x}}_{gcn,k}^m = \mathrm{ReLU}\left(\tilde{\mathbf{A}}_k^m \check{\mathbf{x}}_k^m\right), \tag{3}$$

where $\check{\mathbf{x}}_{gcn,k}^m \in \mathbb{R}^{C_m\times t}$ denotes the feature representation obtained after graph convolution. Notably, according to Eq. (3), the graph convolution operation is performed only along the channel dimension, preserving the temporal information of the EEG signals.

To aggregate intra-region features, we perform average pooling over the valid channels of $\check{\mathbf{x}}_{gcn,k}^m$, producing a compact representation $\check{\mathbf{x}}_{avg,k}^m \in \mathbb{R}^{1\times t}$ for each brain region. If no valid channels are present in a region, the corresponding representation is set to $\mathbf{0} \in \mathbb{R}^{1\times t}$. The average-pooled features $\check{\mathbf{x}}_{avg,k}^m$ from all brain regions are concatenated to form the brain region features $\check{\mathbf{x}}_{avg,k}^{cat} = [\check{\mathbf{x}}_{avg,k}^1; \check{\mathbf{x}}_{avg,k}^2; \dots; \check{\mathbf{x}}_{avg,k}^M] \in \mathbb{R}^{M\times t}$. Considering $K$ views, the final brain region (BR) feature representation is given by $\check{\mathbf{x}}_{BR} = \{\check{\mathbf{x}}_{avg,k}^{cat}\}_{k=1}^K \in \mathbb{R}^{K\times M\times t}$, and is denoted as $\check{\mathbf{x}}_{Pre}^{BR}$ for the pre-training dataset.

To integrate global brain-region spatial information, we apply $K$ convolutional kernels of size $(M, 1)$ to the brain region feature $\check{\mathbf{x}}_{Pre}^{BR}$, yielding the spatially convolved feature representation $\tilde{\mathbf{x}}_{Pre} \in \mathbb{R}^{K\times t}$ . The overall formula for the spatiotemporal extraction block $E_{st}$ is shown below:

$$\tilde{\mathbf{x}}_{Pre} = E_{st}(\mathbf{x}_{Pre}; \theta_{st}), \tag{4}$$

where $\theta_{st}$ represents the neural network parameters of $E_{st}$.

The resulting feature maps $\tilde{\mathbf{x}}_{Pre}$ are then passed to the temporal attention block $E_{attn}$ . Specifically, $\tilde{\mathbf{x}}_{Pre}$ is first rearranged by transposing the convolution channel and temporal dimensions. In this way, the features of all channels at each time point are combined into a token and projected into an embedding space, resulting in the feature representation $\tilde{\mathbf{x}}_{Pre}' \in \mathbb{R}^{t\times d}$, where $t$ denotes the number of tokens and $d$ denotes the embedding dimension of each token. The embedded tokens are then processed by multi-head self-attention (MSA) [34] to capture global temporal dependencies. In this mechanism, the input feature $\tilde{\mathbf{x}}_{Pre}'$ is linearly mapped to three matrices, referred to as query ($\mathbf{Q}$), key ($\mathbf{K}$), and value ($\mathbf{V}$), and the attention scores are computed to weight the values as follows:

$$\mathrm{Attn}(\mathbf{Q},\mathbf{K},\mathbf{V}) = \mathrm{Softmax}\left(\frac{\mathbf{Q}\mathbf{K}^T}{\sqrt{d_k}}\right)\mathbf{V}, \tag{5}$$

where $d_k$ is the dimension of the key vector.

To further enhance the representation capability, we adopt $h$ projection heads to transform the embedding sequences:

$$\mathrm{MSA}(\mathbf{Q},\mathbf{K},\mathbf{V}) = [\mathrm{head}_1, \dots, \mathrm{head}_h], \tag{6}$$
$$\mathrm{head}_l = \mathrm{Attn}(\mathbf{Q}_l, \mathbf{K}_l, \mathbf{V}_l),$$

where $\mathbf{Q}_l, \mathbf{K}_l, \mathbf{V}_l \in \mathbb{R}^{t\times d/h}$ are the projected features for the $l$-th head. Finally, the concatenated output MSA(·) serves as the global temporal representation $\mathbf{h}_{Pre} \in \mathbb{R}^{t\times d}$, capturing both long-range dependencies and diverse feature information of the EEG sequence. The overall formula for the temporal attention block $E_{attn}$ is shown below:

$$\mathbf{h}_{Pre} = E_{attn}(\tilde{\mathbf{x}}_{Pre}; \theta_{attn}), \tag{7}$$

where $\mathbf{h}_{Pre}$ is the global temporal representation.

*2) JEPA-Based Generative Learning Module:* Inspired by prior works [25], we incorporate JEPA into the MGCRL model to better learn global temporal dependencies in EEG signals. JEPA focuses on predicting structured targets in a latent space rather than reconstructing signal-level representations, which encourages the model to capture semantic relationships across different temporal segments. By predicting the embeddings of masked local representations from global temporal context, the MGCRL model can learn more robust and discriminative temporal features within each EEG sample.

The JEPA module consists of three main components: target encoder, context encoder, and predictor. The target encoder module does not participate in gradient descent. The outputs of this module serve as the targets for our generative learning tasks, with its model parameters derived from the context encoder. The context encoder module shares the same architecture as the target encoder module, but it is trainable via gradient descent. Furthermore, it uses the masked features of the input embedding to generate representations for the visible blocks. The predictor utilizes the outputs of the context network and randomly selected masked tokens to regress the targets, which correspond to the outputs of the target network. Below, we detail the processes of creating the targets, the context, and the predictor for training tasks, and provide explanations for the parameterization of these networks.

The target encoder and context encoder are the same network model, consisting of two encoding functions $E_{st}$ and $E_{attn}$. We denote the context encoder parameters as $\theta = (\theta_{st}, \theta_{attn})$. The target encoder parameters are updated at each iteration via an Exponential Moving Average (EMA) of the context encoder parameters. The formula is as follows:

$$\bar{\theta} \leftarrow \eta\bar{\theta} + (1-\eta)\theta, \tag{8}$$

where $\bar{\theta} = (\bar{\theta}_{st}, \bar{\theta}_{attn})$ are parameters for two blocks, and $\eta$ is the hyperparameter that controls the target encoder to receive parameter updates from the context encoder. In this way, when we input $\mathbf{x}_{Pre}$ into the target encoder, we can obtain the global temporal representation $\bar{\mathbf{h}}_{Pre}$:

$$\bar{\mathbf{h}}_{Pre} = E_{attn}(E_{st}(\mathbf{x}_{Pre}; \bar{\theta}_{st}); \bar{\theta}_{attn}). \tag{9}$$

In the context encoder, the input $\mathbf{x}_{Pre}$ is first processed by $E_{st}$ to obtain the temporal features $\tilde{\mathbf{x}}_{Pre}$. We then randomly generate a mask vector $\mathbf{v}_{mask}$ that specifies consecutive temporal positions to be masked in $\tilde{\mathbf{x}}_{Pre}$, yielding the masked temporal features $\hat{\mathbf{x}}_{Pre}$ , with $t_{mask}$ denoting the number of remaining time points after masking. These masked temporal features are subsequently processed by $E_{attn}$, and the overall context encoding process can be formulated as follows:

$$\hat{\mathbf{h}}_{Pre} = E_{attn}(S_{mask}(E_{st}(\mathbf{x}_{Pre}; \theta_{st}), \mathbf{v}_{mask}); \theta_{attn}), \tag{10}$$

where $S_{mask}(\cdot,\cdot)$ denotes the masking operator, and $\hat{\mathbf{h}}_{Pre} \in \mathbb{R}^{t_{mask}\times d}$ represents the global temporal features corresponding to the unmasked positions.

In the predictor, following [25], we employ a transformer encoder, denoted as $g_{pred}$ with parameter $\theta_{pred}$. In addition, we generate a prediction matrix $\mathbf{M}_{pred} \in \mathbb{R}^{t_{pred}\times d}$, that is parameterized by learnable vectors with an added positional embedding. The total sequence length satisfies $t \geq t_{mask} + t_{pred}$, where $t$ is the original sequence length prior to masking. The predictor $g_{pred}$ takes the output of the context encoder $\hat{\mathbf{h}}_{Pre}$ and the prediction matrix $\mathbf{M}_{pred}$ as inputs, producing the corresponding masked global temporal representation $\mathbf{o}_{Pre} \in \mathbb{R}^{t_{pred}\times d}$:

$$\mathbf{o}_{Pre} = g_{pred}\left(\hat{\mathbf{h}}_{Pre}, \mathbf{M}_{pred}; \theta_{pred}\right). \tag{11}$$

During pre-training, the model is optimized by minimizing the L2-norm loss over the predicted embeddings and the ground-truth embeddings. This ensures that the generated predictive embeddings not only retain the key features of the original signal but also exhibit strong contextual consistency. For a batch of $B$ samples, the generative learning loss $\mathcal{L}_2$ is defined as:

$$\mathcal{L}_2 = \frac{1}{B\times t_{pred}}\sum_{i=1}^{B}\sum_{q=1}^{t_{pred}} ||\bar{\mathbf{h}}_{Pre}^{(i)}(\Omega_i[q]) - \mathbf{o}_{Pre}^{(i)}(q)||_2^2, \tag{12}$$

where $\Omega_i[q]$ denotes the original token index of the $q$-th prediction target in the $i$-th sample, and $\bar{\mathbf{h}}_{Pre}^{(i)}(\Omega_i[q])$ and $\mathbf{o}_{Pre}^{(i)}(q)$ denote the corresponding target representation and predicted representation, respectively. By minimizing this loss, the model gradually aligns the predicted embeddings with the target representations, thereby learning the implicit temporal dependencies within the EEG signals.

*3) Masked Dynamic Contrastive Learning Module:* Since the JEPA architecture focuses on generating information within individual samples during training, it is inherently constrained in its ability to model relationships across samples. Therefore, the proposed MGCRL model combines contrastive learning to explicitly focus on modeling inter-sample relationships and improving feature representation. In the proposed MGCRL model, masked and unmasked features are employed as two complementary views in the contrastive learning framework, and their alignment in the feature space allows the model to capture more complete information, improving robustness and generalization.

Following the SimCLR architecture [26], a shared extractor is employed for contrastive learning. The original EEG signal $\mathbf{x}_{Pre}^{(i)}$ is fed into the Region-aware Spatiotemporal Encoder, which yields the unmasked global temporal representation $\mathbf{h}_{Pre}^{(i)}$. Additionally, a neural network $g_{con}$, composed of two non-linear layers with model parameters $\theta_{con}$, is introduced to project the representation into a space suitable for contrastive learning:

$$\mathbf{z}_{Pre}^{(i)} = g_{con}\big(\mathcal{F}_{ap}(\mathbf{h}_{Pre}^{(i)}); \theta_{con}\big), \tag{13}$$

where $\mathcal{F}_{ap}$ represents average pooling and $\mathbf{z}_{Pre}^{(i)} \in \mathbb{R}^{1\times r}$ denotes the mapped features, with $r$ being its dimensionality. On the other hand, we use masked features as additional features for contrastive learning, and their corresponding features can be calculated as follows:

$$\hat{\mathbf{z}}_{Pre}^{(i)} = g_{con}\big(\mathcal{F}_{ap}(\hat{\mathbf{h}}_{Pre}^{(i)}); \theta_{con}\big). \tag{14}$$

Here, $\mathbf{z}_{Pre}^{(i)}$ and $\hat{\mathbf{z}}_{Pre}^{(i)}$ are derived from two different views of the same sample $i$. Given a batch of $B$ samples, we include both views of each sample and construct the anchor set as $\left\{\tilde{\mathbf{z}}_{Pre}^{(j)}\right\}_{j=1}^{2B}$, where $\tilde{\mathbf{z}}_{Pre}^{(j)} = \mathbf{z}_{Pre}^{(j)}$ for $j = 1,2,\dots,B$ and $\tilde{\mathbf{z}}_{Pre}^{(j)} = \hat{\mathbf{z}}_{Pre}^{(j-B)}$ for $j = B+1, B+2, \dots, 2B$. In this way, $\tilde{\mathbf{z}}_{Pre}^{(j)}$ and $\tilde{\mathbf{z}}_{Pre}^{(j+B)}$ $(j = 1,2,\dots,B)$ correspond to two different views of the same sample and form a strong positive pair. Accordingly, the similarity term for the strong positive pairs of anchor $j$ is defined as:

$$\mathrm{S}_{str}^{(j)} = \begin{cases} \exp\left(\frac{sim(\tilde{\mathbf{z}}_{Pre}^{(j)}, \tilde{\mathbf{z}}_{Pre}^{(j+B)})}{\tau}\right), & j = 1,2,\dots,B, \\ \exp\left(\frac{sim(\tilde{\mathbf{z}}_{Pre}^{(j)}, \tilde{\mathbf{z}}_{Pre}^{(j-B)})}{\tau}\right), & j = B+1, B+2 \dots, 2B, \end{cases} \tag{15}$$

where $sim(\cdot,\cdot)$ denotes the cosine similarity, and $\tau$ denotes a temperature parameter. This strong positive term captures the consistency between two different views of the same EEG sample.

In addition to view-based positive pairs, we assume that EEG data elicited by the same stimulus (e.g., watching the same video clip) exhibits stable emotional features over time and should be considered as positive samples in contrastive learning. Furthermore, EEG responses to the same stimulus across different subjects are expected to have similar emotional representations, making them suitable as positive samples to explore the invariant features of different subjects. However, since this similarity is a fuzzy association, we do not treat it as a strong positive but instead use dynamic relationships to measure its strength.

To facilitate this, we introduce a mask matrix based on stimulus label, $\mathbf{M}_{stim} = \{\mathrm{M}_{stim}^{j,k}\}_{j,k=1}^{2B} \in \mathbb{R}^{2B\times 2B}$. Each element of the matrix is defined as:

$$\mathrm{M}_{stim}^{j,k} = \begin{cases} 1, & if\ j \neq k\ and\ s_j = s_k, \\ 0, & otherwise, \end{cases} \tag{16}$$

where $s_j$ and $s_k$ denote the stimulus labels of sample $j$ and $k$, respectively. The diagonal elements $(j = k)$ are excluded to prevent a sample from being considered as a positive pair with itself. In other words, if two distinct samples correspond to the same stimulus (e.g., watching the same video clip), the corresponding element is 1; otherwise, it is 0.

To model the similarity among different subjects exposed to the same stimulus, we introduce a dynamic weighting mechanism. This mechanism computes cosine similarity based on the extracted feature representations $\left\{\tilde{\mathbf{z}}_{Pre}^{(j)}\right\}_{j=1}^{2B}$ and normalizes the values to the range $[0, 1]$, forming a relation matrix $\mathbf{D}_{weights} = \{\mathrm{D}_{weights}^{j,k}\}_{j,k=1}^{2B} \in \mathbb{R}^{2B\times 2B}$. Since the similarity is directly determined by the output of the feature encoder, the model progressively learns more consistent subject-invariant representations during training, leading to re-estimation of the relation matrix at each iteration. As a result, the model continuously adjusts inter-sample weights, enabling adaptive modeling and iterative refinement of cross-subject structural relationships. The similarity term for the dynamic-weighted positive pairs of anchor $j$ is defined as:

$$\mathrm{S}_{dyn}^{(j)} = \frac{1}{\sum_{k=1}^{2B} \mathrm{M}_{stim}^{j,k}} \sum_{k=1}^{2B} \mathrm{M}_{stim}^{j,k} \cdot \mathrm{D}_{weights}^{j,k} \cdot \exp\left(\frac{sim(\tilde{\mathbf{z}}_{Pre}^{(j)}, \tilde{\mathbf{z}}_{Pre}^{(k)})}{\tau}\right). \tag{17}$$

Here, the dynamic-weighted positive term enables adaptive weighting of similarity across samples.

The similarity term for negative pairs of anchor $j$ is defined as:

$$\mathrm{S}_{neg}^{(j)} = \sum_{k=1}^{2B} \mathbb{1}_{[k\neq j]}(1 - \mathrm{M}_{stim}^{j,k}) \cdot \exp\left(\frac{sim(\tilde{\mathbf{z}}_{Pre}^{(j)}, \tilde{\mathbf{z}}_{Pre}^{(k)})}{\tau}\right), \tag{18}$$

where $\mathbb{1}_{[k\neq j]} \in \{0,1\}$ is an indicator function that equals 1 iff $k \neq j$. Finally, the masked dynamic contrastive learning loss $\mathcal{L}_{contrast}$ of a minibatch is computed as:

$$\mathcal{L}_{contrast} = \frac{1}{2B} \sum_{j=1}^{2B} \left(-\log\left(\frac{\mathrm{S}_{str}^{(j)} + \mathrm{S}_{dyn}^{(j)}}{\mathrm{S}_{neg}^{(j)}}\right)\right). \tag{19}$$

This loss jointly considers intra-sample consistency (i.e., different views of the same EEG sample) and inter-sample relationships (i.e., samples under the same stimulus) to learn discriminative representations.

In our proposed model, the total loss of the self-supervised framework integrates the generative learning module and the contrastive learning module, which respectively capture intra-sample information and inter-sample information. This forms a natural complementarity, significantly enhancing unsupervised learning. The specific formula is as follows:

$$\mathcal{L} = \mathcal{L}_2 + \mathcal{L}_{contrast}. \tag{20}$$

Here, the generative and contrastive branches are complementary and jointly optimized through direct integration.

### *D. Fine-tuning and Test*

In the fine-tuning stage, we use a limited set of labeled data $(\mathbf{X}_F^s, \mathbf{Y}_F^s)$ from a specific dataset s. At this stage, following the JEPA [25], we remove the generative learning module to simplify the model architecture. Meanwhile, inspired by Core-tuning [35], we retain the contrastive learning module to maintain feature discriminability and improve emotion recognition performance. Specifically, the target encoder and predictor in the JEPA-based generative learning module are removed after pre-training. The context encoder is retained as the backbone feature extractor, and its masked-feature representations are directly used in contrastive learning. In addition, a classification head $g_{cls}$ is introduced, and the model is supervised using a standard cross-entropy loss $\mathcal{L}_{cls}$ to optimize classification performance. The classification head is dataset-specific and is adapted to match the number of emotion categories in each target dataset, thereby naturally handling label space differences across datasets.

During the test stage, we use an unseen subject whose acquisition device and classification task match those of the fine-tuning dataset $s$, denoted as $\mathbf{X}_T^s$, to assess the model's generalization ability across different subjects. Algorithm 1 summarizes the proposed MGCRL framework.

**Algorithm 1** Masked Generative-Contrastive Representation Learning

**Input**: Unlabeled pre-training data $\mathbf{X}_{Pre} = \{\mathbf{x}_{Pre}^{(i)}\}_{i=1}^{N_{Pre}}$; fine-tuning data $\mathbf{X}_F^s = \{\mathbf{x}_F^{(i)}\}_{i=1}^{N_F}$ with labels $\mathbf{Y}_F^s = \{\mathbf{y}_F^{(i)}\}_{i=1}^{N_F}$; test data $\mathbf{X}_T^s = \{\mathbf{x}_T^{(i)}\}_{i=1}^{N_T}$; number of epochs $Epoch$ ; batch size $B$.
**Model**: The spatiotemporal extraction block $E_{st}$; the temporal attention block $E_{attn}$; the predictor $g_{pred}$; the nonlinear projection head $g_{con}$; the classification head $g_{cls}$.
**Output**: Trained model parameters and predicted labels $\hat{\mathbf{Y}}_T^s$.

**Stage 1: Generalized Pre-training**

```
Initialize model parameters θ_st, θ_attn, θ_pred, and θ_con, and set θ̄_st ← θ_st, θ̄_attn ← θ_attn.
For e = 1: Epoch do
  For each mini-batch X^B_Pre sampled from X_Pre do
    For each sample x^(i)_Pre in X^B_Pre do
      Feed x^(i)_Pre separately into the region-aware spatiotemporal encoder, target encoder, and context encoder to obtain the global temporal representations h^(i)_Pre, h̄^(i)_Pre, and ĥ^(i)_Pre, respectively, based on Eqs. (4), (7), (9), and (10).
      Feed ĥ^(i)_Pre and a learnable prediction matrix M_pred into the predictor to predict the masked global temporal representation o^(i)_Pre based on Eq. (11).
      Feed h^(i)_Pre and ĥ^(i)_Pre into the projection head to obtain features z^(i)_Pre and ẑ^(i)_Pre, respectively, based on Eq. (13) and Eq. (14).
    End For
    Use {h̄^(i)_Pre, o^(i)_Pre}^B_{i=1} to calculate the generative learning loss L_2 based on Eq. (12).
    Use {z^(i)_Pre, ẑ^(i)_Pre}^B_{i=1} to construct the anchor set {z̃^(j)_Pre}^{2B}_{j=1}.
    Construct the stimulus mask matrix M_stim based on Eq. (16) and re-estimate the relation matrix D_weights from the current batch representations.
    Construct the strong positive pairs {S^(j)_str}^{2B}_{j=1}, dynamic-weighted positive pairs {S^(j)_dyn}^{2B}_{j=1} and negative pairs {S^(j)_neg}^{2B}_{j=1} using {z̃^(j)_Pre}^{2B}_{j=1}, M_stim, and D_weights, as specified in Eqs. (15), (17) and (18).
    Use {S^(j)_str}^{2B}_{j=1}, {S^(j)_dyn}^{2B}_{j=1}, and {S^(j)_neg}^{2B}_{j=1} to calculate the masked dynamic contrastive learning loss L_contrast based on Eq. (19).
    Use L_2 and L_contrast to calculate the overall pre-training loss L based on Eq. (20).
    Update {θ_st, θ_attn, θ_pred, θ_con} by minimizing the loss L.
    Update {θ̄_st, θ̄_attn} via EMA based on Eq. (8).
  End For
End For
```

**Stage 2: Fine-tuning**

```
Load model parameters θ_st, θ_attn, and θ_con.
Initialize the classification head g_cls with parameters θ_cls.
For e = 1: Epoch do
  For each mini-batch X^B_F sampled from X^s_F do
    For each sample x^(i)_F in X^B_F do
      Feed x^(i)_F separately into the region-aware spatiotemporal encoder and the context encoder to obtain the global temporal representations h^(i)_F and ĥ^(i)_F, respectively, based on Eqs. (4), (7) and (10).
      Feed h^(i)_F into the classification head g_cls to predict the label ŷ^(i)_F.
      Feed h^(i)_F and ĥ^(i)_F into the projection head to obtain features z^(i)_F and ẑ^(i)_F, respectively, based on Eq. (13) and Eq. (14).
    End For
    Use {y^(i)_F, ŷ^(i)_F}^B_{i=1} to calculate the cross-entropy loss L_cls.
    Use {z^(i)_F, ẑ^(i)_F}^B_{i=1} to construct the anchor set {z̃^(j)_F}^{2B}_{j=1}, construct M_stim, re-estimate D_weights from the current batch, and calculate the masked dynamic contrastive learning loss L_contrast based on Eqs. (15)-(19).
    Use L_cls and L_contrast to calculate the overall fine-tuning loss L_fine
    Update {θ_st, θ_attn, θ_con, θ_cls} by minimizing the loss L_fine.
  End For
End For
```

**Stage 3: Test**

```
Use the fine-tuned model to predict emotion labels Ŷ^s_T.
```

## IV. EXPERIMENTS

In this section, we introduce the adopted datasets and present the implementation details of the proposed method. We then compare MGCRL with several state-of-the-art self-supervised learning methods and conduct ablation studies to evaluate the contributions of key components. The effects of fine-tuning data scale are further analyzed, followed by sensitivity and computational complexity analyses. In addition, brain region visualizations are provided to explicitly illustrate the contributions of different brain regions. Finally, the generalization ability of the model is rigorously evaluated under cross-session cross-subject settings.

### *A. Datasets*

The proposed model is evaluated on several widely used affective EEG datasets, including FACED [36], SEED-IV [37], SEED-V [38], and SEED-VII [39].

The FACED dataset consists of 32-channel EEG data collected from 123 subjects (48 males and 75 females) watching 28 emotion-elicitation video clips. Excluding the reference electrodes A1 and A2 leaves 30 effective EEG channels. EEG signals were recorded at a sampling rate of 250Hz. The experiment covers nine emotion categories, including amusement, inspiration, joy, tenderness, anger, fear, disgust, sadness, and neutral.

The SEED-IV dataset contains 62-channel EEG data from 15 subjects (7 males and 8 females). Each subject participates in three sessions on different days, with 24 trials per session. EEG signals were recorded at a sampling rate of 1000Hz. The dataset covers four emotion categories: happy, sad, fear, and neutral.

The SEED-V dataset was collected using the same equipment as SEED-IV and includes EEG recordings from 16 subjects (6 males and 10 females). Each subject participated in three sessions on different days, with 15 trials per session. EEG signals were recorded at a sampling rate of 1000Hz. The dataset covers five emotion categories: happy, sad, neutral, fear, and disgust.

The SEED-VII dataset comprises 62-channel EEG data from 20 subjects (10 males and 10 females). Each subject participated in four sessions on different days, with 20 trials per session. EEG signals were recorded at a sampling rate of 1000Hz. The dataset covers seven emotion categories: happy, sad, fear, disgust, neutral, anger, and surprise. Each session involved only five out of the seven emotions, which helped reduce the impact of subjects switching into too many emotional states within a single experiment.

TABLE I
LEAVE-ONE-SUBJECT-OUT EMOTION RECOGNITION ACCURACY ON SEED-IV, SEED-V, AND SEED-VII DATASETS (ACC. AND STD. %).

| Model | Datasets | | |
|---|---|---|---|
| | FACED → SEED-IV | FACED→ SEED-V | FACED→ SEED-VII |
| BIOT [18] | 31.68±2.73 | 27.39±2.95 | 28.77±3.98 |
| TS-TCC [14] | 36.63±4.41 | 28.78±3.74 | 28.87±4.10 |
| EEG2Rep [19] | 37.20±5.71 | 30.25±6.16 | 31.03±5.58 |
| LaBraM [16] | 36.63±3.80 | 31.98±4.59 | 30.52±4.06 |
| EEGPT [41] | 35.27±3.96 | 29.39±4.16 | 30.63±3.98 |
| CBraMod [42] | 33.27±3.08 | 26.64±2.87 | 29.83±3.62 |
| **MGCRL (Ours)** | **41.47±7.37** | **34.99±7.40** | **34.73±6.20** |

### *B. Implementation Details*

We apply a unified preprocessing pipeline to all datasets. First, the last 30 seconds of each video clip are selected to capture the most intense emotional responses [36], and the corresponding EEG signals are divided into non-overlapping 1-second segments, with each segment treated as an independent sample. Second, the EEG signals are downsampled to 200 Hz to handle inconsistent sampling rates across datasets. Third, following prior work [40], external noise is reduced using a Butterworth filter, including a 48–52 Hz band-stop filter and a 0.5–45 Hz band-pass filter. Finally, each segment is normalized using the Z-score method.

The model architecture and optimization settings are kept consistent across all experiments. Following [31], the number of temporal convolutions (i.e., $K$ ) is set to 40. In line with JEPA [25], the target block scale is randomly selected from the range (0.15, 0.20) and the aspect ratio in the range (0.75, 1.5), while the context block scale is randomly selected from the range (0.85, 1.0). Moreover, the encoder dimension is set to 128, the Transformer depth is set to 3, and the number of multi-head self-attention heads is set to 8. MGCRL is optimized using SGD, with both pre-training and fine-tuning conducted for 100 epochs and a batch size of 64. The learning rate is set to 0.01 during pre-training and 0.0001 during fine-tuning.

### *C. Method Comparisons*

To evaluate the effectiveness of the proposed MGCRL model, its performance is compared with several representative cross-dataset learning methods, including BIOT [18], TS-TCC [14], EEG2Rep [19], LaBraM [16], EEGPT [41], and CBraMod [42]. All baseline methods are implemented under the experimental settings reported in their original papers to ensure a fair comparison. Additionally, we further optimize learning rates and batch sizes to allow each baseline method to achieve its best performance. In the following experiments, within-session leave-one-subject-out cross-validation (LOSO-CV) is employed to evaluate cross-subject generalization performance of different methods.

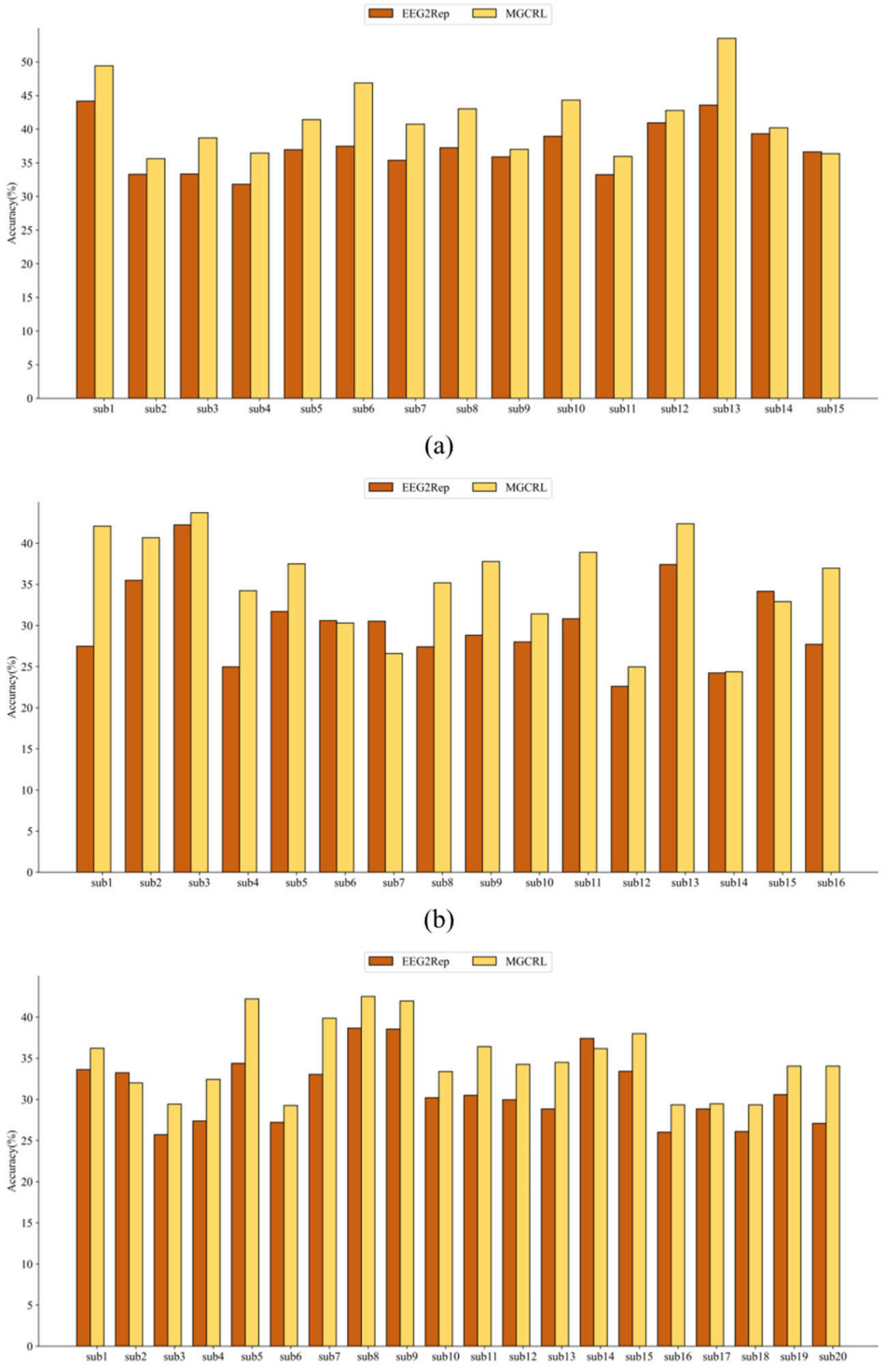

(c)

Fig. 4. Experimental results of MGCRL and EEG2Rep on three datasets for each subject. (a) SEED-IV. (b) SEED-V. (c) SEED-VII.

Specifically, for each target dataset (SEED-IV, SEED-V, and SEED-VII), subjects are partitioned according to LOSO, where one subject is held out for testing while the remaining subjects are used for training. The results are first averaged across all subjects within each session, and then averaged across all sessions to determine the final emotion recognition accuracy. This evaluation protocol directly reflects the model's ability to generalize to unseen subjects, which is essential for real-world EEG-based emotion recognition applications. All methods are pre-trained on the FACED dataset and fine-tuned on the SEED-IV, SEED-V, and SEED-VII datasets. The emotion recognition accuracies (± standard deviation) are summarized in Table I, from which we can make the following observations.

First, TS-TCC has demonstrated high efficiency on time-series data with limited labeled data, but its effectiveness decreases on EEG signals, which exhibit complex spatial dependencies. This limitation arises because TS-TCC lacks mechanisms to model spatial dependencies. In contrast, the proposed MGCRL model explicitly captures both spatial and temporal relationships in EEG data, demonstrating superior performance and generalization ability on each dataset.

Second, LaBraM, BIOT, EEGPT, and CBraMod learn by reconstructing signal-level representations. However, this approach is susceptible to noise during training, potentially introducing irrelevant or noisy information into the generated features. By contrast, MGCRL employs a JEPA-based generative learning mechanism that effectively suppresses noise while better capturing fine-grained neural dynamics and distinguishing subtle emotional variations.

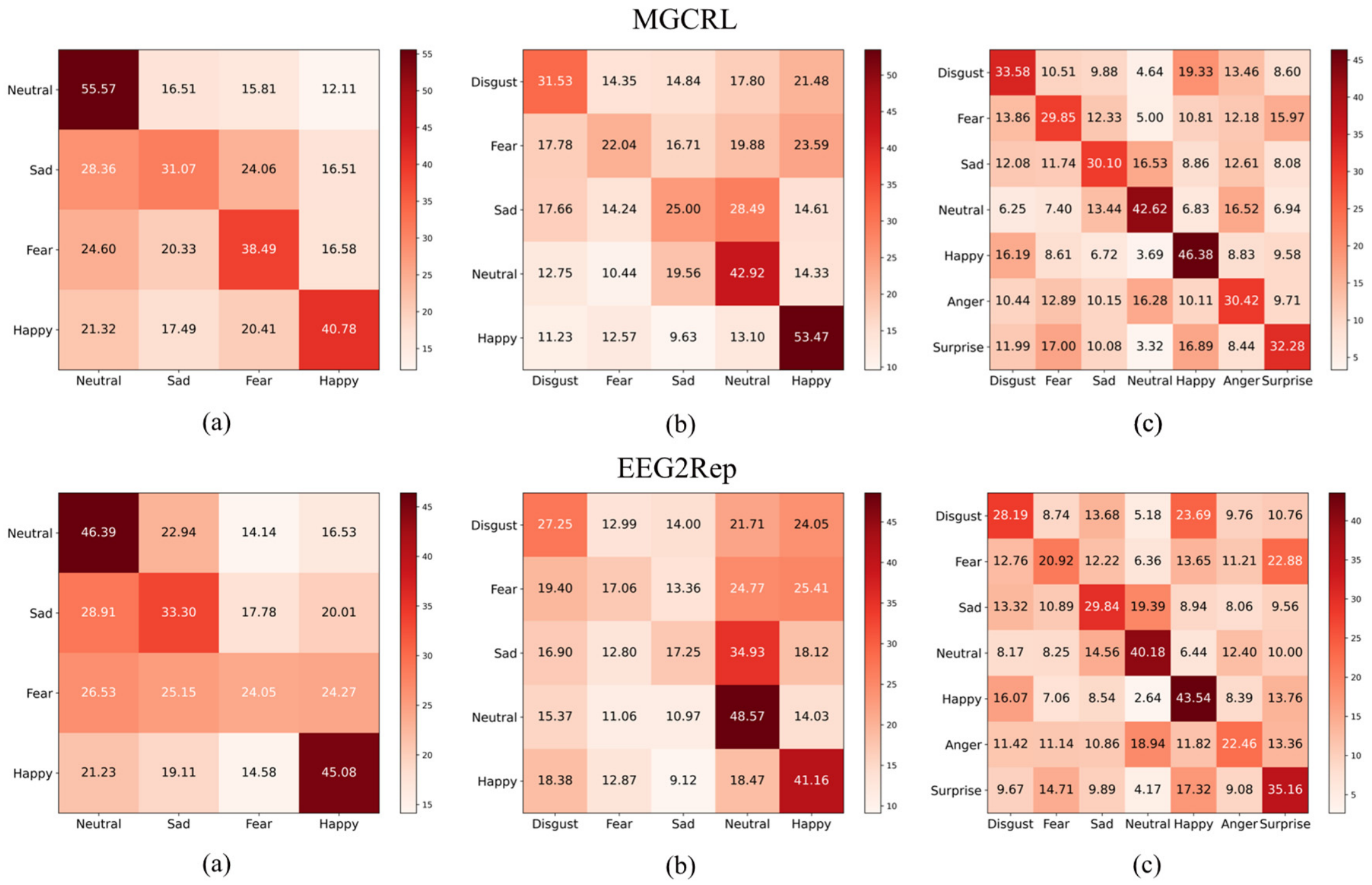

Fig. 5. Confusion matrices of MGCRL and EEG2Rep on three datasets. (a) SEED-IV. (b) SEED-V. (c) SEED-VII.

Third, although EEG2Rep employs a JEPA-based generative learning mechanism, it does not fully account for individual differences in emotional responses and does not effectively model cross-subject variability and the semantic similarities between emotional states. In contrast, our model employs masked dynamic contrastive learning, fully leveraging the temporal continuity of emotions and the cross-subject similarity of emotional states, thereby simultaneously enhancing feature discriminability and the model's cross-subject generalization ability.

To evaluate the robustness of MGCRL, we compare subject-wise accuracy on the SEED-IV, SEED-V, and SEED-VII datasets with that of EEG2Rep, the baseline model with the best overall performance. For each subject, results are averaged across all sessions. As shown in Fig. 4, MGCRL achieves higher recognition accuracy than EEG2Rep for the vast majority of subjects across all three datasets, demonstrating strong robustness.

To further assess the model's ability to discriminate between different emotional categories, we plot the confusion matrices of MGCRL and EEG2Rep on the SEED-IV, SEED-V, and SEED-VII datasets. As shown in Fig. 5, a consistent trend can be observed across the three datasets. Both methods achieve relatively high recognition accuracy for neutral and happy emotions, whereas performance for negative emotions is generally lower. Compared with EEG2Rep, MGCRL achieves superior accuracy on the majority of categories and demonstrates more stable and robust overall performance.

TABLE II
RESULTS OF ABLATION STUDY IN MGCRL ON SEED-IV, SEED-V, AND SEED-VII DATASETS (ACC. AND STD. %).

| Model | Datasets | | |
|---|---|---|---|
| | FACED → SEED-IV | FACED → SEED-V | FACED → SEED-VII |
| **MGCRL (Ours)** | **41.47±7.37** | **34.99±7.40** | **34.73±6.20** |
| w/o $\mathcal{L}_2$ | 40.95±7.05 | 34.45±7.40 | 34.59±6.00 |
| w/o IBR-GCN | 39.90±6.89 | 32.99±7.87 | 33.59±6.02 |
| w/o $\mathcal{L}_{contrast}$ | 37.87±5.86 | 30.68±6.76 | 27.80±4.17 |
| w/o $\Delta_k^m$ | 41.12±7.19 | 34.42±7.18 | 34.02±6.13 |

## D. Ablation Study

*1) The importance of different modules:* To further analyze the impact of each key component in the MGCRL model on the final performance, we conduct ablation experiments by removing individual modules. In each experiment, only one specific component is removed while all other components remain unchanged. The experiments are as follows:

1) *w/o* $\mathcal{L}_2$: We remove the generative learning loss $\mathcal{L}_2$ to evaluate the model's ability to characterize subtle emotional states.
2) *w/o IBR-GCN*: We remove the IBR-GCN to assess its impact on the spatial relationships.
3) *w/o* $\mathcal{L}_{contrast}$ : We remove the masked dynamic contrastive learning loss $\mathcal{L}_{contrast}$ to evaluate the model's discriminative ability and generalization capability enhanced through emotional characteristics.
4) w/o $\Delta_k^m$: We remove the $\Delta_k^m$ in Eq. (1) to evaluate its contribution to supplementing connectivity information.

TABLE III
PERFORMANCE COMPARISON OF DIFFERENT CONTRASTIVE LOSSES ON SEED-IV, SEED-V, AND SEED-VII DATASETS (ACC. AND STD. %).

| Model | Datasets | | |
|---|---|---|---|
| | FACED → SEED-IV | FACED → SEED-V | FACED → SEED-VII |
| OCLL | 34.82±3.21 | 28.56±3.58 | 28.12±3.62 |
| DCLL | 37.10±4.73 | 28.28±2.75 | 29.78±4.08 |
| **MDCLL(ours)** | **41.47±7.37** | **34.99±7.40** | **34.73±6.20** |

TABLE IV
PERFORMANCE OF DIFFERENT TARGET BLOCK RANGES ON SEED-IV, SEED-V, AND SEED-VII (ACC. AND STD. %).

| Target Block Ranges | Datasets | | |
|---|---|---|---|
| | FACED → SEED-IV | FACED → SEED-V | FACED → SEED-VII |
| (0.10, 0.20) | 40.78±7.54 | 34.64±7.26 | 34.57±6.14 |
| **(0.15, 0.20)** | **41.47±7.37** | **34.99±7.40** | 34.73±6.20 |
| (0.20, 0.25) | 41.00±7.38 | 34.65±7.39 | **35.12±5.96** |

The results of these evaluations are shown in Table II, demonstrating that all components of MGCRL are both significant and indispensable. Removing the $\mathcal{L}_2$ causes a slight performance drop, suggesting its role in capturing subtle emotional states. Similarly, removing the IBR-GCN leads to a decrease in performance, highlighting the importance of modeling spatial dependencies among EEG channels within individual brain regions. Furthermore, removing the $\Delta_k^m$ results in a slight performance decrease, indicating its effectiveness in providing global guidance to supplement the connectivity information. Finally, removing the $\mathcal{L}_{contrast}$ leads to a significant decline in the model's ability to distinguish between samples, underscoring its critical contribution to improving discriminative performance.

*2) The effectiveness of improved contrastive learning loss:* To evaluate the effectiveness of the proposed masked dynamic contrastive learning strategy in Eq. (19), we compare the performance of the original contrastive learning loss (OCLL) [26] and masked dynamic contrastive learning loss (MDCLL). In addition, we remove the masked features from MDCLL to obtain dynamic contrastive learning loss (DCLL), in order to demonstrate the role of masked features as complementary views. Specifically, OCLL corresponds to the standard contrastive learning loss based on conventional data augmentations (i.e., Gaussian noise and temporal jittering). DCLL performs contrastive learning based on unmasked features, constructing positive pairs using the sample itself and samples under the same stimulus condition, and further introduces a dynamic weighting mechanism. MDCLL is the masked dynamic contrastive learning mechanism introduced in our framework.

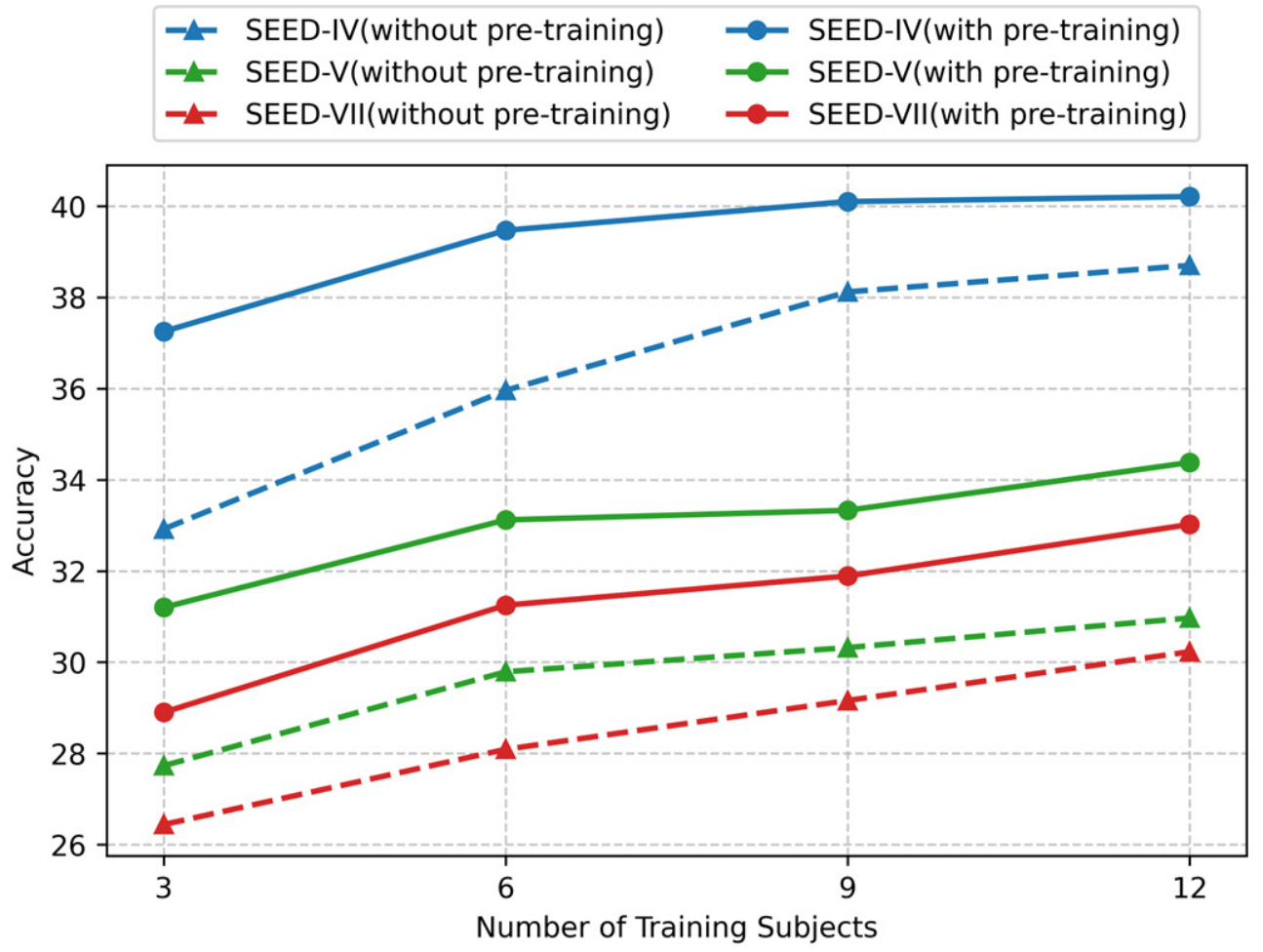

Fig. 6. The results of fine-tuning different numbers of training subjects on SEED-IV, SEED-V, and SEED-VII.

As shown in Table III, OCLL yields relatively low recognition accuracy, suggesting that conventional data augmentation strategies tend to disrupt informative patterns in EEG signals. DCLL provides only limited and inconsistent performance improvements, indicating that modeling dynamic inter-sample relationships alone may be insufficient for learning robust discriminative representations. In contrast, MDCLL achieves substantially better performance by introducing masked features as complementary views, enabling the model to jointly capture intra-sample consistency and cross-subject emotional similarity.

## *E. Fine-Tuning Data Scale Impact*

To further evaluate the effectiveness of the proposed model, we fine-tune it on the SEED-IV, SEED-V, and SEED-VII datasets using different numbers of training subjects, and the corresponding results are reported in Fig. 6. It can be observed that the model's performance consistently improves as the number of training subjects increases. Moreover, across all datasets, models initialized with pre-training weights consistently outperform those trained from scratch. This indicates that the pre-training strategy provides a strong prior representation, facilitating more efficient fine-tuning and yielding superior performance even in settings with a limited number of training subjects. Notably, the pretrained model fine-tuned with data from only 6 subjects on SEED-IV and SEED-VII, and only 3 subjects on SEED-V, outperforms the model without pre-training that uses data from 12 subjects, indicating that pre-training effectively reduces the reliance on labeled data.

TABLE V
COMPUTATIONAL COMPLEXITY COMPARISON IN TERMS OF PARAMETERS AND FLOPS.

| Model | Parameters | FLOPs |
|---|---|---|
| BIOT [18] | 3.19M | 0.256G |
| TS-TCC [14] | 0.13M | 0.009G |
| EEG2Rep [19] | 0.09M | 0.004G |
| LaBraM [16] | 5.82M | 0.563G |
| EEGPT [41] | 0.11M | 0.024G |
| CBraMod [42] | 21.77M | 0.336G |
| **MGCRL(Ours)** | 1.29M | 0.041G |

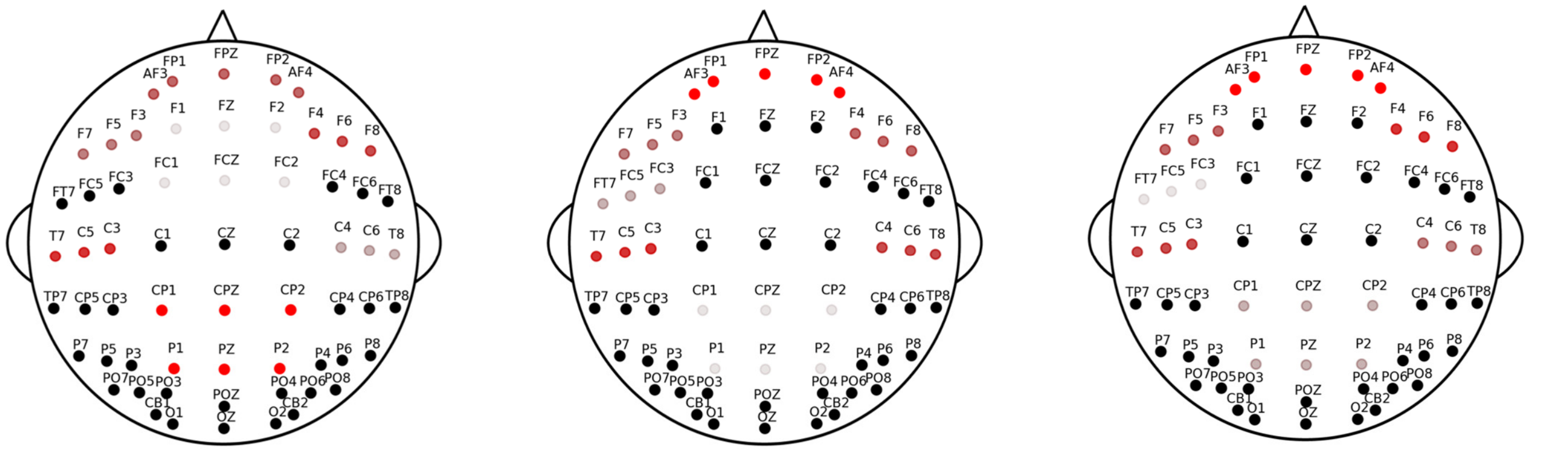

(a) (b) (c)

Fig. 7. The brain region weights are visualized across the three datasets, with the top seven most important regions highlighted in red. Variations in color intensity and transparency indicate the relative weight of each region. (a) SEED-IV. (b) SEED-V. (c) SEED-VII.

### *F. Sensitivity Analysis*

To further investigate the influence of the masking strategy, we conduct a sensitivity analysis on the target block scale within the JEPA-based generative learning module. Since predictive representation learning relies on a sufficiently informative and spatially distributed context block, the context block scale is fixed to (0.85, 1.0) throughout the experiments, while different target block scales are evaluated. Specifically, we investigate three target block ranges: (0.10, 0.20), (0.15, 0.20), and (0.20, 0.25), as summarized in Table IV. Experimental results on the SEED-IV, SEED-V, and SEED-VII datasets show that the target block range of (0.15, 0.20) achieves the best average performance across the datasets.

We observe that excessively small prediction targets may make the prediction task overly easy, thereby limiting the model's ability to learn sufficient semantic information. In contrast, overly large prediction regions may make the prediction task excessively difficult, reducing the effectiveness of contextual predictive learning for non-stationary EEG signals. These results suggest that a moderate predictive masking scale provides a better balance between prediction difficulty and representation quality.

TABLE VI
LEAVE-ONE-SUBJECT-OUT EMOTION RECOGNITION ACCURACY ACROSS SESSIONS ON SEED-IV AND SEED-V DATASETS (ACC. AND STD. %).

| Model | Datasets | |
|---|---|---|
| | FACED→ SEED-IV | FACED→SEED-V |
| BIOT [18] | 31.28±2.74 | 26.46±2.66 |
| TS-TCC [14] | 34.29±4.08 | 27.00±3.20 |
| EEG2Rep [19] | 35.70±5.71 | 30.14±5.96 |
| LaBraM [16] | 34.67±3.78 | 30.56±4.62 |
| EEGPT [41] | 33.95±3.71 | 28.05±3.73 |
| CBraMod [42] | 32.69±3.44 | 26.26±2.32 |
| **MGCRL(Ours)** | **40.07±6.89** | **33.56±6.61** |

Note: Cross-session evaluation is not applicable to SEED-VII, as the emotion categories are not consistent across sessions.

### *G. Computational Complexity Analysis*

We further analyze the computational complexity of MGCRL in terms of parameter count and FLOPs and compare it with existing EEG representation learning models. As shown in Table V, MGCRL maintains a moderate computational cost among the compared methods. In addition, compared with large-scale models such as BIOT, LaBraM, and CBraMod, MGCRL requires substantially fewer computational resources while still maintaining competitive performance. These results indicate that MGCRL achieves a favorable balance between computational efficiency and representation effectiveness, suggesting its potential applicability for EEG-based emotion recognition.

### *H. Interpretability*

To quantify the contribution of each brain region in the learning task, we examine the $K$ convolutional kernel weights of size $(M, 1)$ learned by the spatial convolution over brain regions. Specifically, for each brain region, we average the absolute values of its weights across the $K$ kernels to obtain a spatial importance score. We then rank the $M$ brain regions according to their scores and select the seven regions with the highest scores as the key brain regions. As shown in Fig. 7, these regions are highlighted in red, with variations in color intensity and transparency employed to indicate the relative importance of different brain regions. Notably, these key brain regions exhibit a common distribution pattern across different datasets, which aligns with the existing research findings [43], [44], [45]. The identified regions also agree with established neuroscientific evidence: the frontal lobe dominates emotional regulation and conscious thinking, and the temporal lobe is responsible for processing auditory information related to emotion and memory [46]. In addition, previous studies have further demonstrated that the frontal, temporal, and parietal regions play crucial roles in emotion processing evoked by audiovisual stimuli [47], [48]. These observations suggest that the spatial convolution module may highlight brain regions potentially relevant to emotion processing across the examined datasets.

### I. Across-Session Generalization

To evaluate the generalization ability of MGCRL, we adopt cross-session leave-one-subject-out cross-validation (LOSO-CV), which reflects a more practical scenario and evaluates the model's capability to generalize to novel emotion-elicitation stimuli (i.e., unseen videos). As shown in Table VI, our proposed MGCRL model achieves the highest accuracy under the cross-session cross-subject setting, reaching 40.07% on SEED-IV and 33.56% on SEED-V. These results demonstrate the superior robustness and generalization ability of our model across different sessions compared with baseline methods. Moreover, the performance drop compared with the within-session cross-subject setting (i.e., Table I) is relatively small. This indicates that, even under unseen video stimuli, MGCRL is still able to learn more stable representations of emotional states rather than overly relying on specific video stimuli, thereby demonstrating strong cross-session generalization capability.

## V. Conclusion

This paper proposes a pre-training framework that integrates generative and contrastive learning to address challenges arising from device differences, label incompatibility, and channel mismatches. We design a region-aware spatiotemporal encoder to capture local spatial and functional dependencies, and introduce masked dynamic contrastive learning to enhance discriminability across emotion categories and generalization across subjects. Compared with the strongest self-supervised baselines, MGCRL improves emotion recognition accuracy by 4.27%, 3.01%, and 3.70% in the within-session cross-subject setting on SEED-IV, SEED-V, and SEED-VII, respectively. In the cross-session cross-subject setting, it improves accuracy by 4.37% and 3.00% on SEED-IV and SEED-V, respectively.

In summary, MGCRL significantly improves EEG-based emotion recognition by integrating contrastive and generative paradigms in a mutually complementary manner. MGCRL provides a general and flexible framework that can be extended to other EEG-based representation learning tasks. Nevertheless, the current framework relies on stimulus-related information for contrastive pair construction, which may limit its applicability to datasets lacking such information. Future research will explore more general self-supervised learning strategies.